\documentclass[12pt,a4paper]{article}
\usepackage[a4paper,left=2.0cm,right=2.0cm,top=2.5cm,bottom=2.5cm]{geometry}

\usepackage{url}
\usepackage{hyperref}

\usepackage{float}

\usepackage{algorithm}	
\usepackage[noend]{algpseudocode}
\usepackage{amsmath}
\usepackage{amssymb}
\usepackage[title]{appendix}
\usepackage{adjustbox,lipsum}
\usepackage{threeparttable}
\usepackage{graphicx}
\usepackage{centernot}

\algdef{SE}[SUBALG]{Indent}{EndIndent}{}{\algorithmicend\ }%
\algtext*{Indent}
\algtext*{EndIndent}

\newcommand{\bigO}{\mathcal{O}}

\begin{document}
\date{}
\title{Imputing missing values with unsupervised random trees}
\author{David Cortes}
\maketitle

\begin{abstract}
This work proposes a non-iterative strategy for missing value imputations which is guided by similarity between observations, but instead of explicitly determining distances or nearest neighbors, it assigns observations to overlapping buckets through recursive semi-random hyperplane cuts, in which weighted averages are determined as imputations for each variable. The quality of these imputations is oftentimes not as good as that of chained equations, but the proposed technique is much faster, non-iterative, can make imputations on new data without re-calculating anything, and scales easily to large and high-dimensional datasets, providing a significant boost over simple mean/median imputation in regression and classification metrics with imputed values when other methods are not feasible.
\end{abstract}

\section{Introduction}
When designing statistical models from tabular data for supervised learning tasks such as regression or classification, oftentimes it happens that some of the observations available for fitting such models are missing values in one or more variables, usually due to reasons such as poor data collection practices, loss of information, participants dropping out of a survey, or similar. Many methods such as \cite{cart} or \cite{xgb} overcome this issue by using heuristics to handle missing information - decision tree methods in particular, due to their splitting nature that takes one variable at a time, are particularly well suited for implicit handling of missing data without a-priori imputation (\cite{nas}), but other methods such as generalized linear models or support vector machines cannot handle missing values in the same way, and when using them on a dataset with missing entries, these entries have to either be dropped or imputed.

Typical strategies for imputing the missing entries include: replacing them with the column mean or median, determining the most similar observations (nearest neighbors) according to the non-missing variables and taking a simple or weighted average of the missing variable(s) from them (\cite{kuhn}), producing a latent representation of the data by some low-rank matrix factorization that minimizes errors on the non-missing entries and from which the missing entries are then reconstructed (\cite{koren}), and iterative imputation that starts with some basic imputation for all values and then cycles through each variable by constructing a model to predict the missing values from the non-missing observations, replacing the earlier imputation with the model prediction and repeating until convergence (\cite{mice}, \cite{missforest}).

This work proposes a strategy for imputation of missing values in data based on unsupervised decision trees similar to the Isolation Forest (\cite{iso}) algorithm or to Random Tree Embeddings\footnote{\url{https://scikit-learn.org/stable/modules/generated/sklearn.ensemble.RandomTreesEmbedding.html}}, which as opposed to common decision tree algorithms such as \textsc{C5.0} (\cite{c45}) or \textsc{CART} (\cite{cart}), do not aim at predicting one variable from a set of other variables, but rather aim at grouping observations according to random criteria.

The general idea is to construct multiple and independent binary trees by recursively splitting the data according to some condition, with each terminal node in a tree producing imputations for every variable according to the non-missing observations that end in that node. At prediction time, observations are passed down these trees until reaching a terminal node, from which the imputations are taken and averaged across the terminal nodes reached in each tree. The data is split through hyperplane cuts similar to \cite{sci}, but using a pooled information gain criterion like in \cite{cart} instead, which results in improved imputations compared to averaged gain and fully-random cuts. Missing values are temporarily median-imputed at each non-terminal node in order to continue passing observations down the tree, but these imputations are not used for the final calculations.

\section{Fair-cut trees}

The Isolation Forest algorithm (\cite{iso}) - initially proposed for outlier detection - explored the idea of producing decision trees that split (assign to different branches of a binary tree node) a set of multi-dimensional points/observations according to whether they are greater or smaller than a threshold chosen uniformly at random within the range of a randomly-determined variable/feature, continuing the process recursively on the subset of observations that fall into each branch until observations/points become isolated (put alone in one branch), following the logic that outlier observations will become isolated with fewer splits than expected. Under an alternative view, this same procedure manages to group observations that are closer in the feature space according to the variables that were chosen, with nodes further down the tree grouping together observations that are more and more similar, and outliers alternatively being interpreted as the ones that are dissimilar from all other observations - a scaled version of the (negative) average depth at which two observations are separated can be shown to constitute a distance metric with many desirable properties as analyzed in \cite{dist}.

As such, when building a random tree like this, the nearest neighbors of an observation are more likely to be those that end in the same terminal node if there is a pre-determined height limit, or those with which a larger number of ancestors is shared. The random tree procedure, however, at no point implies explicitly computing pairwise distances or even defining a distance metric - the splitting logic of random trees, when averaged across multiple runs, will implicitly put together similar observations more often according to the sample data distribution, being invariant to the scales of variables and taking complex relationships between variables into account (see \cite{dist}). It is also a computationally efficient procedure in large and/or high-dimensional datasets as the procedure scales only linearly with the sample size, depth, and number of trees ($\bigO(ndt)$) when growing the trees, and only linearly with the number of trees and their depth when passing observations down the tree ($\bigO(dt)$).

There are some logical improvements that can make this random tree procedure better at grouping similar observations (and isolating outliers faster in the process). As shown in \cite{ext}, splitting by only one variable at a time introduces some biases which can be overcomed by using hyperplane cuts instead - that is, the split criterion is based on a threshold set on a linear combination of randomly-chosen variables instead of a threshold on a single variable’s value at a time.

The uniformly-random threshold can also be changed to a guided threshold as proposed in \cite{sci}, by trying to minimize the standard deviations of the obtained linear combination when calculated on the observations within each branch ($(\sigma - \frac{\sigma_{\text{left}} + \sigma_{\text{right}} }{2}) / \sigma$), which manages to isolate clustered minority observations faster. This concept is similar to the information gain criterion used by supervised decision tree algorithms such as \textsc{CART} or \textsc{C5.0}, but with the target variable being split according to a threshold on itself instead of being split according to some other variable.

If the objective is to group similar observations instead of separating clustered outliers, it makes more sense to maximize a weighted or pooled average for the gain criterion that could take into account the number of observations that are being assigned to each branch ($(\sigma - \frac{n_{\text{left}} \sigma_{\text{left}}  +  n_{\text{right}} \sigma_{\text{right}} }{n}) / \sigma$), in the same way as \textsc{CART} for example. A pooled gain-maximizing split criterion, when applied on the same variable being split, tends to favor more “even” or “fair” splits - for example, a symmetric distribution would get split in half for each branch, and a gamma distribution would get the long tail assigned to a branch, whereas an averaged gain criterion will put the maximum or minimum valued observation alone in one branch. This produces a more fair split, in the sense that it will prefer a half-each division unless there is some reasonable cluster(s) with reasonable size that could be assigned to a branch as minority. As well, the criterion has a natural threshold that can tell if the groups are being made more homogeneous: if the gain is negative, then the split is not making the observations under each branch more similar to each other, and this threshold can be increased if desired. There is unfortunately no fast way of determining the linear combination that would have the split point with the highest gain, so different linear combinations can be tried and only the best (highest-gain) among them taken, as proposed in \cite{sci}.

\begin{figure}[tph!]
\centerline{\includegraphics[totalheight=5cm]{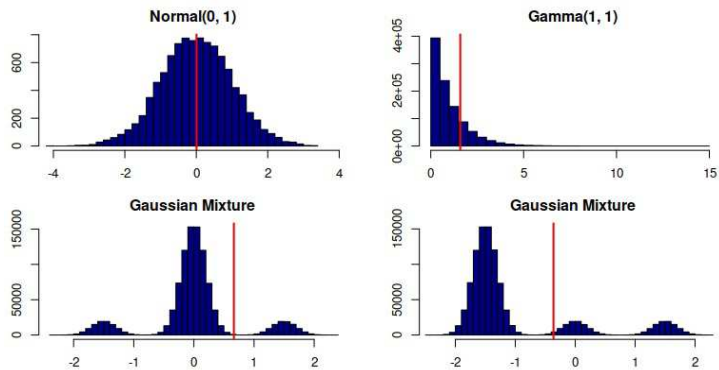}}
    \caption{Split point that maximizes pooled gain on randomly-generated data when splitting a variable by a threshold on itself}
    \label{fig:verticalcell}
\end{figure}

An obvious shortcoming of this procedure is that it doesn’t handle missing values, and missing values cannot be handled by assigning a whole observation to both branches simultaneously with decreased weights on them, since there is more than one variable deciding the split and there could be missing values in multiple variables or even no complete observations. An alternative for handling missing values is to resort to mean or median imputation - if these are re-calculated at each tree node, then their quality should get better as the depth increases.

The logic can also be extended to handle categorical variables by assigning a random coefficient in the linear combination for each possible category, thereby mixing them into the gain calculation along with the numeric variables. If the numeric variables are standardized, and the coefficients for numeric columns and categorical values are both drawn from e.g. a standard normal distribution, then the variance in numeric variables will tend to be higher, which is perhaps preferable in many situations, whereas a wider range of possible coefficients for categorical values (e.g. the product of two draws from a normal distribution) would make them more comparable.

The procedure, named hereafter as \textsc{FairCutTree}, is illustrated below - the steps are the same as the \textsc{iTreeExtEnh} proposed in \cite{dist}, but with the addition of the gain criterion.

\begin{algorithm}[H]
\caption{\textsc{FairCutTree}}\label{FairCutTree}
\hspace*{\algorithmicindent} \textbf{Inputs} $\mathbf{X}_{n,p}$ (input data points), $d_{\text{max}}$ (max depth), $d$ (current depth), $m$ (number of splitting dimensions), $\tau$ (number of trials), $g_{\text{min}}$ (minimum gain for a split) \\
\hspace*{\algorithmicindent} \textbf{Output} Tree node with left branch $T_l$, right branch $T_r$, subset of variables $\textsc{u}$, chosen coefficients $\textsc{z}$, temporary imputation values $\textsc{r}$, split point $q$
\begin{algorithmic}[1]
\If {$|\{\mathbf{x} \in \mathbf{X}\}| = 1$ or $d = d_{\text{max}}$}
	\State Terminate procedure (return empty output $\emptyset$)
\Else
	\State Initialize $g_{\text{best}} = -\infty$
	\For {trial $1..\tau$}
		\State Initialize linear combination $\mathbf{y} := \mathbf{0}$ for each point in $\mathbf{X}$
		\State Initialize empty set of coefficients $z := \emptyset$
		\State Choose a subset $\textsc{u}$ of $m$ variables at random from $1..p$ such that $\mathbf{X}_{:,v}, v \in \textsc{u}$ has at least 2 different values (fewer than $m$ if not possible, terminate if none has at least 2 different values)
		\For {each variable $v \in \textsc{u}$}
			\State Draw a random coefficient $z_v \sim \text{Normal}(0, 1)$
			\State Standardize coefficient as $z_v := \frac{z_v}{\sigma_{\mathbf{X}_{:,v}}}$
			\State Add coefficient to set $\textsc{z} := \textsc{z} \cup z_v$
			\State Determine temporary imputation value as $r_v = \text{Median}\{x_{i,v} \:\: \text{known} \}$
			\State Add imputation value to the set $\textsc{r} := \textsc{r} \cup r_v$
			\State Update $y_i := y_i + z_v 
				\begin{cases}
				x_{i,v} & x_{i,v} \:\:\text{known} \\
				r_v & x_{i,v} \:\:\text{unknown}
			 	\end{cases} \forall x_{i,v}$
		\EndFor
		
		\State Determine split point $q_{\tau} = \text{argmax}_q (\sigma^y - \frac{n_l \sigma_l^y + n_r \sigma_r^y }{n}) / \sigma^y \:\:\:\: \text{s.t.} \:\: l = \{ y_i | y_i \leq q \}, \: r = \{ y_i | y_i > q \}$
		\State Set trial gain as $g_{\tau} = (\sigma^y - \frac{n_l \sigma_l^y + n_r \sigma_r^y }{n}) / \sigma^y \:\:| \:\:\:\: l = \{ y_i | y_i \leq q_{\tau} \}, \: r = \{ y_i | y_i > q_{\tau} \}$
		\If {$g_{\tau} > g_{\text{best}}$}
			\State Update $g_{\text{best}} := g_{\tau}$
			\State Set $\textsc{z}_{\text{best}} = \textsc{z}$, $\textsc{r}_{\text{best}} = \textsc{r}$, $\mathbf{y}^{\text{best}} = \mathbf{y}$, $q_{\text{best}} = q_{\tau}$
		\EndIf
	\EndFor

	\If {$g_{\text{best}} < g_{\text{min}}$}
		\State Terminate procedure (return empty output $\emptyset$)
	\EndIf
	\State Determine subsets $\mathbf{X}_l = \{ {\mathbf{x}_i \in \mathbf{X} | y_i^{\text{best}} \leq q_{\text{best}}} \}$, $\mathbf{X}_r = \{ {\mathbf{x}_j \in \mathbf{X} | y_j^{\text{best}} > q_{\text{best}}} \}$.
	\State \Return tree node with left branch $T_l = \textsc{FairCutTree}(\mathbf{X}_l, d, d + 1, m, \tau, g_{\text{min}})$, right branch $T_r = \textsc{FairCutTree}(\mathbf{X}_r, d, d + 1, m, \tau, g_{\text{min}})$, subset of variables $\textsc{u}_{\text{best}}$, coefficients $\textsc{z}_{\text{best}}$, temporary imputation values $\textsc{R}_{\text{best}}$, split point $q_{\text{best}}$
\EndIf
\end{algorithmic}
\end{algorithm}

(For a software implementation of the procedure, an equivalent but more numerically-stable formulation might center the variables beforehand by subtracting their mean, setting the threshold on a linear combination of the standardized values instead)

\section{Node imputations}

Using the notion that the \textsc{FairCutTree} is grouping together more similar observations with each split, potential imputation values could be determined at every node in the tree according to the observations that reach that node and which have no missing values, or according to observations that were separated the least splits ago if some variable has only-missing observations. Intuitively, the further down the tree (away from the root) a node is, the higher the impact it should have in an imputation (e.g. if an observation is isolated at the first split, new observations falling into that same node are potentially not too similar to that observation), and the fewer observations there are in a deep node, the more likely it is that they are nearest neighbors.

This work suggests a weighted-average imputation that looks only at the terminal node in which an observation falls under each tree in an ensemble of \textsc{FairCutTree}s (hereafter \textsc{FairCutForest}), and weights the imputations of these terminal nodes according to their depth and the number of observations they had at training (tree growing) time. In order to have more reasonable results, the trees need not be grown to full-depth (this is similar to setting the number of neighbors in kNN). There is also no necessity for sub-sampling or bootstrapping the sample observations from which a tree is grown.

The procedure for determining imputation values from an ensemble (forest) of \textsc{FairCutTrees} is outlined below:

\begin{algorithm}[H]
\caption{\textsc{NodeImputation}}\label{NodeImputation}
\hspace*{\algorithmicindent} \textbf{Inputs} $T$, (Node of a \textsc{FairCutTree}), $\mathbf{X}_{n,p}$ (points passed to $T$), $d$ (depth of $T$), $T_{\text{parent}}$ (parent node), $n_{\text{min}}$ (minimum non-missing obs.)  \\
\hspace*{\algorithmicindent} \textbf{Output} Numeric imputation values $\hat{\textbf{x}}_{p}$, imputation weights $\mathbf{w}_p$
\begin{algorithmic}[1]
\For {each variable $v$}
	\State Determine number of known observations $k = |\{ x_{i,v} \:\:\text{known} \}|$
	\State Set imputation value $\hat{x}_v = 
	\begin{cases}
	\frac{\sum_{\text{known}} x_{i,v}}{k} & \text{if} \:\: k \geq n_{\text{min}} \\
	\hat{x}_v \in I_{T_{\text{parent}}} & \text{otherwise}
	\end{cases}
	$
	\State Set imputation weight $w_v = 
	\begin{cases}
	\frac{d}{\sqrt{k}} & \text{if} \:\: k \geq n_{\text{min}} \\
	\frac{d}{2 \sqrt{n}} & \text{otherwise}
	\end{cases}
	$
\EndFor
\State Set imputation values $I_T = \{ \hat{\mathbf{x}}, \mathbf{w} \}$ for node $T$
\State \Return $I_T$
\end{algorithmic}
\end{algorithm}

\begin{algorithm}[H]
\caption{\textsc{FairCutForest} imputation}\label{FairCutForest imputation}
\hspace*{\algorithmicindent} \textbf{Inputs} $F_t$ (ensemble of $t$ \textsc{FairCutTree}s), $\mathbf{x}_p$ (point with missing values) \\
\hspace*{\algorithmicindent} \textbf{Output} Point with imputed vales $\widetilde{\mathbf{x}}_p$
\begin{algorithmic}[1]
\State Initialize imputations $\widetilde{\mathbf{x}}_{p} = \mathbf{0}$
\State Initialize weights $\mathbf{w}_p = \mathbf{0}$
\For {each tree $\tau$ in $F$}
	\State Pass $\mathbf{x}$ down the tree until reaching terminal node $T$
	\For {each variable $v$ missing in $\mathbf{x}$}
		\State Update imputations with node info $\widetilde{x}_v := \widetilde{x}_v + \widetilde{x}_v^{I_T} w_v^{I_T} $
		\State Update weights with node info $w_v := w_v + w_v^{I_T}$
	\EndFor
\EndFor
\For {each variable $v$ missing in $\mathbf{x}$}
	\State Impute $x_v := \frac{\widetilde{x}_v}{w_v}$
\EndFor
\State \Return $\mathbf{x}$
\end{algorithmic}
\end{algorithm}

The recommended hyperparameters are $\tau = 20$, $m = 3$, $d_{\text{max}} = 3 \lceil \log_2 n \rceil$, $n_{\text{min}} = 3$, $n_{\text{trees}} = 500$, $g_{\text{min}} = 0$. Increasing $\tau$ translates into a small increase in the quality of the imputations, while the optimal $d_{\text{max}}$ and $n_{\text{min}}$ varies in each dataset. Weighting the number of observations differently in the imputation (e.g. not using the number of observations in a terminal node) sometimes also results in improved results, especially when $d_{\text{max}}$ is low. For faster imputation times, $\tau$ and $d_{\text{max}}$ can be reduced, and the trees can be fit to sub-sampled data.

Notice however that this imputation technique, just like kNN or similar methods, does not take into account patterns of non-randomly-missing variables (e.g. censored data), nor does it enforce constraints such as some variable being calculated deterministically from other variables (e.g. $\text{BMI} = \frac{\text{Weight}}{\text{Height}^2}$), or some variable only allowed to take certain values according to a different variable (e.g. "pregnant" and "gender").

\section{Comparison against other methods}

The \textsc{FairCutForest} imputation method proposed here (the implementation used is made open source and freely available\footnote{\url{https://www.github.com/david-cortes/isotree}}) was compared against simple median imputation, kNN imputation from the software library SciKit-Learn (\cite{sklearn}), matrix factorization from software library "cmfrec" (\cite{cmfrec}), and chained equations (hereafter referred to interchangeably as "iterative imputation") from the software libraries "missForest" (\cite{missforest}), "mice" (\cite{mice}), and SciKit-Learn, using datasets of different sizes and characteristics from which some values were set randomly as missing, or which originally came with many missing values. The imputed data was then given as input to regression and classification algorithms which included linear and logistic regression from SciKit-Learn, random forest (\cite{rf}) from software "randomForest" (\cite{rfcran}), and gradient-boosted decision trees from software "XGBoost" (\cite{xgb}) and "CatBoost" (\cite{catboost}). Evaluation metrics were calculated using k-fold cross-validation, and the timing was calculated for a single run in the full data for both imputation-model fitting and producing imputations from the model on new data (which in this case was the same training data). The benchmarks, depending on the dataset, were run either on a 3.2GHz AMD Ryzen 2700 processor with 8 cores, or on a 2.0GHz Intel Xeon Knight's Landing processor with 64 virtual CPUs from Google Cloud Platform. All timings make use of the full available number of threads, and all methods are used with their default hyperparameters in each software, except for the method proposed here.

For \textsc{FairCutForest}, three combinations of hyperparameters were compared:
\itemize{
	\item "Large", with the previously recommended hyperparameters ($\tau = 20$, $m = 3$, $d_{\text{max}} = 3 \lceil \log_2 n \rceil$, $n_{\text{min}} = 3$, $n_{\text{trees}} = 500$, $g_{\text{min}} = 0$).
	\item "Mid", with fewer and shallower trees (same depth recommended for outlier detection), evaluating fewer splits ($\tau = 10$, $m = 3$, $d_{\text{max}} = \lceil \log_2 n \rceil$, $n_{\text{min}} = 3$, $n_{\text{trees}} = 100$, $g_{\text{min}} = 0$).
	\item "Small", with mostly the same hyperparameters as "Mid" but growing the trees with sub-samples of 5,000 points each instead of taking the full data, and using a maximum depth that reflects the reduced sample size ($d_{\text{max}} = 13$).
}

\subsection{California housing prices dataset}

This regression dataset\footnote{\url{https://scikit-learn.org/stable/modules/generated/sklearn.datasets.fetch_california_housing.html}} (20,640 rows, 8 columns) contains basic geographic and demographic information about districts in California (all of them numeric columns), with the suggested target variable being the median house value in the district.

Given the low dimensionality and the type of variables, this is the kind of problem in which iterative imputation is most appropriate and in which it tends to outperform other methods. Another reasonable alternative here is kNN, but making predicions on new data from it is too slow for a real-time system.

In order to evaluate imputation quality, some of the entries were set randomly as missing across rows/columns, and a situation in which every row has 1 missing column (at random) was also evaluated. Another scenario in which only 10\% of the rows were used was also simulated in order to have a better idea of what happens when the problem size varies.

The evaluation metric used was root mean squared error (RMSE), which is evaluated through 10-fold cross-validation. The timings are measured on a 2.0GHz Intel Xeon Knight's Landing processor with 64 virtual CPUs from Google Cloud Platform.

No feature engineering or filtering of rows was performed, but for kNN, the variables were standardized by subtracting their mean and dividing them by their standard deviation.

\begin{table}[H]
\caption {California housing prices prediction}
\begin{adjustbox}{max width=\textwidth}{\centering
\begin{tabular}{|r|c|c|c|c|c|c|c|}
 \hline
   &   & \multicolumn{5}{|c|}{ \textbf{RMSE}} \\
 \textbf{Imputation} & \textbf{Regressor} & \textbf{1 NA/row} & \textbf{10\% NA} & \textbf{10\% NA, 1\% obs.} & \textbf{15\% NA} & \textbf{20\% NA} \\
 \hline
\textsc{FairCutForest} (large) & Linear & 0.830 & 0.816 & 0.862 & 0.849 & 0.873 \\
\textsc{FairCutForest} (mid) & Linear & 0.857 & 0.845 & 0.831 & 0.868 & 0.888 \\
\textsc{FairCutForest} (small) & Linear & 0.859 & 0.847 & - & 0.870 & 0.889 \\
\textsc{kNN} & Linear & 0.806 & 0.815 & \textbf{0.71} & 0.859 & 0.901 \\
Iterative (Linear Reg.) & Linear & 0.821 & 0.785 & 0.844 & 0.843 & 0.877 \\
Iterative (XGBoost) & Linear & \textbf{0.775} & \textbf{0.770} & 0.868 & \textbf{0.788} & \textbf{0.809} \\
Median & Linear & 0.868 & 0.849 & 0.860 & 0.882 & 0.906 \\
\cline{1-7}
\textsc{FairCutForest} (large) & XGBoost & 0.738 & 0.729 & 0.781 & 0.755 & 0.776 \\
\textsc{FairCutForest} (mid) & XGBoost & 0.745 & 0.730 & 0.801 & 0.755 & 0.784 \\
\textsc{FairCutForest} (small) & XGBoost & 0.743 & 0.735 & - & 0.759 & 0.786 \\
\textsc{kNN} & XGBoost & 0.721 & 0.726 & 0.783 & 0.768 & 0.811 \\
Iterative (Linear Reg.) & XGBoost & 0.706 & 0.699 & 0.844 & 0.727 & 0.758 \\
Iterative (XGBoost) & XGBoost & \textbf{0.691} & \textbf{0.690} & \textbf{0.774} & \textbf{0.712} & \textbf{0.734} \\
Median & XGBoost & 0.731 & 0.731 & 0.764 & 0.755 & 0.774 \\
Implicit & XGBoost & 0.725 & 0.717 & 0.783 & 0.740 & 0.734 \\
 \hline
\end{tabular}}\end{adjustbox}
\begin{tablenotes}
      \footnotesize \item[*](RMSE measured through 10-fold cross-validation, with full data producing RMSE of 0.742 for linear regression, and 0.638 for XGBoost.)
\end{tablenotes}
\end{table}

\begin{table}[H]
\caption {California housing prices imputation times}
\begin{adjustbox}{max width=\textwidth}{\centering
\begin{tabular}{|r|c|c|c|c|c|c|}
 \hline
   &   & \multicolumn{4}{|c|}{ \textbf{Time (s)}} \\
 \textbf{Imputation} & \textbf{Task} & \textbf{1 NA/row} & \textbf{10\% NA} & \textbf{15\% NA} & \textbf{20\% NA} \\
 \hline
\textsc{FairCutForest} (large) & Build imputer & 7.28 & 4.00 & 3.96 & 3.98 \\
\textsc{FairCutForest} (mid) & Build imputer & 2.89 & 1.47 & 1.54 & 1.48 \\
\textsc{FairCutForest} (small) & Build imputer & 1.6 & 1.57 & 1.57 & 1.62 \\
Iterative (Linear Reg.) & Build imputer & 0.54 & 0.29 & 0.27 & 0.49 \\
Iterative (XGBoost) & Build imputer & 22.98 & 25.82 & 24.87 & 24.10 \\
\cline{1-6}
\textsc{FairCutForest} (large) & Impute data & 0.24 & 0.16 & 0.20 & 0.23 \\
\textsc{FairCutForest} (mid) & Impute data & 0.07 & 0.04 & 0.05 & 0.06 \\
\textsc{FairCutForest} (small) & Impute data & 0.06 & 0.04 & 0.05 & 0.05 \\
\textsc{kNN} & Impute data & 36.35 & 21.80 & 32.98 & 36.21 \\
Iterative (Linear Reg.) & Impute data & 0.06 & 0.04 & 0.04 & 0.08 \\
Iterative (XGBoost) & Impute data & 0.18 & 0.15 & 0.19 & 0.24 \\
 \hline
\end{tabular}}\end{adjustbox}
\begin{tablenotes}
      \footnotesize \item[*](Timings measured on a 2.0GHz Intel Xeon Knight's Landing processor with 64 virtual CPUs from Google Cloud Platform. All timings make use of the full available number of threads.)
\end{tablenotes}
\end{table}

\subsection{Cover type dataset}

This classification dataset\footnote{\url{https://archive.ics.uci.edu/ml/datasets/covertype}} (581,012 rows, 54 columns, 7 classes) contains a mixture of numeric and binary variables about cartographic measurements and properties of soil plots, with the target variable of interest being forest cover type, and missing values introduced in the same way as for the previous dataset.

Iterative imputation and kNN do not scale well to these problem scales, so the data was sub-sampled by taking only 10\% of the observations, with an additional test taking only 1\% of the observations, and kNN was not evaluated for cross-validation metrics due to the long times it takes. While most of the columns are binary, these were treated as numeric for imputation purposes due to limitations of the software and methods being compared against, without any post-processing of these imputations. The linear regression used here for imputations uses $l_2$ regularization on its coefficients.

The classifiers used multinomial loss minimization as objective, and the metric evaluated was classification accuracy. This can however be a rather variable metric despite the large number of observations, with the imputed data some times producing higher accuracy than the full data with no missing values. Some basic feature transformations such as standardizing variables can easily increase test set accuracy for logistic regression, but this work only evaluated methods with the data as-is.

\begin{table}[H]
\caption {Cover type multi-class classification (sub-sampled)}
\begin{adjustbox}{max width=\textwidth}{\centering
\begin{tabular}{|r|c|c|c|c|c|c|c|}
 \hline
   &   & \multicolumn{5}{|c|}{ \textbf{Classification Accuracy (\%)}} \\
 \textbf{Imputation} & \textbf{Classifier} & \textbf{1 NA/row} & \textbf{10\% NA} & \textbf{10\% NA, 1\% obs.} & \textbf{15\% NA} & \textbf{20\% NA} \\
 \hline
\textsc{FairCutForest} (large) & Logistic & 60.18 & 59.19 & 57.02 & \textbf{58.53} & \textbf{57.56} \\
\textsc{FairCutForest} (mid) & Logistic & 59.99 & 59.16 & 58.07 & 57.76 & 56.68 \\
\textsc{FairCutForest} (small) & Logistic & \textbf{60.38} & \textbf{59.37} & 58.39 & 58.17 & 57.37 \\
Iterative (Linear Reg.) & Logistic & 59.76 & 57.67 & 57.87 & 57.17 & 55.65 \\
Iterative (XGBoost) & Logistic & 59.35 & 58.77 & \textbf{58.73} & 58.44 & 57.50 \\
Median & Logistic & 59.86 & 58.01 & 56.68 & 57.46 & 55.80 \\

\cline{1-7}

\textsc{FairCutForest} (large) & XGBoost & \textbf{66.57} & \textbf{66.50} & 59.79 & \textbf{65.14} & \textbf{63.98} \\
\textsc{FairCutForest} (mid) & XGBoost & 66.32 & 65.86 & 58.07 & 64.146 & 63.18 \\
\textsc{FairCutForest} (small) & XGBoost & 66.03 & 65.52 & 58.93 & 64.92 & 63.25 \\
Iterative (Linear Reg.) & XGBoost & 65.97 & 65.39 & 60.97 & 64.6 & 63.35 \\
Iterative (XGBoost) & XGBoost & 66.40 & 65.04 & \textbf{62.02} & 64.09 & 62.75 \\
Median & XGBoost & 65.85 & 65.74 & 59.79 & 64.94 & 63.82 \\
Implicit & XGBoost & 66.20 & 65.42 & 58.07 & 64.72 & 63.78 \\

 \hline
\end{tabular}}\end{adjustbox}
\begin{tablenotes}
      \footnotesize \item[*](Accuracy measured through 5-fold cross-validation, with full data producing accuracies of 59.95\% for logistic regression, and 66.10\% for XGBoost.)
\end{tablenotes}
\end{table}

\begin{table}[H]
\caption {Cover type imputation times (sub-sampled)}
\begin{adjustbox}{max width=\textwidth}{\centering
\begin{tabular}{|r|c|c|c|c|c|c|}
 \hline
   &   & \multicolumn{4}{|c|}{ \textbf{Time (s)}} \\
 \textbf{Imputation} & \textbf{Task} & \textbf{1 NA/row} & \textbf{10\% NA} & \textbf{15\% NA} & \textbf{20\% NA} \\
 \hline
 
\textsc{FairCutForest} (large) & Build imputer & 32.15 & 36.08 & 35.65 & 38.14 \\
\textsc{FairCutForest} (mid) & Build imputer & 7.99 & 4.58 & 4.79 & 5.03 \\
\textsc{FairCutForest} (small) & Build imputer & 4.31 & 4.57 & 4.78 & 4.54 \\
Iterative (Linear Reg.) & Build imputer & 12.34 & 41.17 & 41.14 & 41.83 \\
Iterative (XGBoost) & Build imputer & 270.26 & 277.7 & 287.3 & 290.78 \\
\cline{1-6}
\textsc{FairCutForest} (large) & Impute data & 0.89 & 1.02 & 1.04 & 1.08 \\
\textsc{FairCutForest} (mid) & Impute data & 0.28 & 0.27 & 0.29 & 0.28 \\
\textsc{FairCutForest} (small) & Impute data & 0.19 & 0.20 & 0.20 & 0.21 \\
\textsc{kNN} & Impute data & 428.38 & 1228.81 & 1565.00 & 1867.97 \\
Iterative (Linear Reg.) & Impute data & 1.2 & 5.39 & 5.91 & 7.42 \\
Iterative (XGBoost) & Impute data & 3.68 & 5.65 & 7.89 & 8.98 \\
 \hline
\end{tabular}}\end{adjustbox}
\begin{tablenotes}
      \footnotesize \item[*](Timings measured on a 2.0GHz Intel Xeon Knight's Landing processor with 64 virtual CPUs from Google Cloud Platform. All timings make use of the full available number of threads.)
\end{tablenotes}
\end{table}

\subsection{Hypothyroid dataset}

This dataset\footnote{\url{https://archive.ics.uci.edu/ml/datasets/Thyroid+Disease}} (3,163 observations, 26 columns) contains a mixture of numeric, boolean, and categorical columns, but unlike the previous ones, the missing values are not simulated as the data already comes with missing values, with some variables being non-randomly missing - for example, it contains columns "sex" and "pregnant", with all males having missing values for "pregnant". As it contains mixed-type columns, not all benchmarked software is able to produce imputations or to use the categorical columns as categorical (e.g. SciKit-Learn requires dummy or one-hot encoding), so the choice of comparisons here was different. Software such as "missForest" and "mice" cannot produce imputations on new data from an already-fitted model, so the timing in this case was for the whole imputation-model fitting and reconstruction of missing values combined.

For \textsc{FairCutForest}, categorical and boolean variables were imputed by following the logic of assigning a random coefficient to each category in such columns, which was drawn from the same standard normal distribution as the coefficients for numeric columns, and categorical imputations decided by proportion of observations in each node having each possible categorical value, with the final imputation decided according to the one presenting the highest weighted proportion.

The suggested target variable for this dataset is "hypothyroid", but this is a rather uninteresting variable to analyze as most models would achieve AUC $> 0.988$ - instead, the data was used for regression by trying to predict "age" among the observations that were not missing it (2,717). Linear models with no feature engineering were not benchmarked for this dataset due to presenting performance too far below tree-based regressors. Models were evaluated by 10-fold cross-validation, with the missing values imputed beforehand for all data (including those without age) instead of building an imputation model with the training data of each fold. These imputations made use of the target variable itself.

Using the extra observations with missing age (the target variable) for imputing other variables managed to significantly improve performance for the \textsc{FairCutForest} model with certain combinations of hyperparameters, which was not the case for "missForest" and "mice". It should be remarked that it is not a problem to use the target variable for imputations on new data with \textsc{FairCutForest}, since it does not need to be known for the imputation model to work.

\begin{table}[H]
\caption {Hypothyroid, age prediction}
\begin{adjustbox}{max width=\textwidth}{\centering
\begin{tabular}{|r|c|c|c|c|}
 \hline
 \textbf{Imputation} & \textbf{RMSE (randomForest)} & \textbf{RMSE (CatBoost)} & \textbf{Time (s) fit-impute} \\
 \hline
\textsc{FairCutForest} (large) (all data) & 18.176 & 17.981 & 14.31 \\
\textsc{FairCutForest} (mid) (all data) & \textbf{15.56} & \textbf{15.926} & 0.86 \\
missForest (all data) & 17.591 & 17.613 & 34.84 \\
mice (all data) & 18.125 & 17.927 & 10.88 \\
Median/Mode (all data) & 18.341 & 18.134 & -  \\
\cline{1-4}
\textsc{FairCutForest} (large) (w/age only) & 18.137 & 17.938 & 11.60 \\
\textsc{FairCutForest} (mid) (w/age only) & 18.066 & 17.964 & 0.73 \\
missForest (w/age only) & \textbf{17.437} & \textbf{17.516} & 23.83 \\
mice (w/age only) & 18.059 & 17.874 & 7.99 \\
Median/Mode (w/age only) & 18.332 & 18.134 & -  \\
 \hline
\end{tabular}}\end{adjustbox}
\begin{tablenotes}
      \footnotesize \item[*](Timings measured on a 3.2GHz AMD Ryzen 2700 processor with 8 cores using all available threads, RMSE measured through 10-fold cross-validation)
\end{tablenotes}
\end{table}

\subsection{MovieLens 100k dataset}
This is a different problem from previous ones - the data consists of movie rating triplets User-Item-Rating (944 users/rows, 1683 items/columns, 100k ratings - see \cite{movielens} for details), with most entries being missing (93.7\%), and the objective is to rank the missing entries for each user so as to recommend movies according to their predicted rating.

The most common evaluation metric for this problem is RMSE in a hold-out sample of triplets (an approach popularized by the Netflix Prize competition), which however is very sensitive to differences that do not change the induced ranking of predictions, so pearson correlation was also measured (it should be noted that this is not a ranking metric either, and is not measured on a per-user basis). Typical regression techniques are not suitable for this problem, but this is where matrix factorization outperforms other approaches. An iterative imputation method would be computationally infeasible for a dataset with these dimensions.

The parameters used for \textsc{FairCutForest} were $m = 100$, $d_{\text{max}} = 100$, $\tau = 25$, $g_{\text{min}} = 0$, $n_{\text{min}} = 3$. The test set consisted of the last $10,000$ ratings in the data rather than k-folds (it's more realistic to predict future behavior based on past behavior), which after discarding users and items that were not in the training set, was reduced to $9,980$ triplets. The timings for matrix factorization use sparse matrices, while the timings for \textsc{FairCutForest} use dense matrices. The imputation timing in the case of matrix factorization is only for the test set triplets, while for \textsc{FairCutForest} includes whole-row imputations.

While the quality of the recommendations from the method proposed here lags behind that from specialized recommender system techniques, it does at least manage to surpass non-personalized recommendations.

\begin{table}[H]
\caption {MovieLens 100k}
\begin{adjustbox}{max width=\textwidth}{\centering
\begin{tabular}{|r|c|c|c|c|}
 \hline
 \textbf{Method} & \textbf{RMSE} & $\rho$ & \textbf{Time fit (s)} & \textbf{Time predict (s)} \\
 \hline
Matrix Factorization & 0.966 & 0.533 & 3.54 & 0.003 \\
\textsc{FairCutForest} & 1.008 & 0.447 & 25.3 & 1.4 \\
Mean imputation & 1.027 & 0.414 & - & - \\
 \hline
\end{tabular}}\end{adjustbox}
\begin{tablenotes}
      \footnotesize \item[*](Timings measured on an AMD Ryzen 2700 3.2GHz processor with 8 cores, using dense matrices for \textsc{FairCutForest} and sparse matrices for Matrix Factorization, RMSE measured on the last 10,000 ratings)
\end{tablenotes}
\end{table}

\section{Conclusions}

This work proposed a missing-value imputation algorithm (\textsc{FairCutForest} imputation) which is based on similarity according to unsupervised semi-random hyperplane splits. The quality of these imputations was evaluated by measuring regression and classification metrics with models fitted to imputed data from both randomly and non-randomly missing entries. 

The method was compared against other imputation strategies and was found to decrease loss metrics for regression and classification with missing data when compared against simple median/mode imputations, in some situations also managing to decrease loss metrics further than kNN and iterative imputation, but is usually behind iterative methods that couple low-bias algorithms for imputation.

The \textsc{FairCutForest} imputation was on the other hand found to be much faster and able to scale to very high-dimensional datasets in which kNN and iterative imputation methods are not computationally feasible, offering a reasonable alternative for low-latency systems and for large and/or high-dimensional datasets.

\bibliographystyle{plain}
\bibliography{imp}

\end{document}